\documentclass{article}
\usepackage{spconf,amsmath,graphicx}
\usepackage{enumitem}
\usepackage{caption}
\usepackage{wrapfig}
\usepackage{afterpage}
\usepackage{placeins}
\usepackage{array}
\usepackage{tabularx}
\usepackage{dblfloatfix}
\usepackage{floatrow}
\newfloatcommand{capbtabbox}{table}[][\FBwidth]

\title{UNSUPERVISED AUDIO-VISUAL SUBSPACE ALIGNMENT \protect\\FOR HIGH-STAKES DECEPTION DETECTION}
\name{Leena Mathur and Maja J Matarić}
  
  \address{Department of Computer Science\\
      University of Southern California, Los Angeles, CA}
\begin{document}
%
\maketitle

\begin{abstract}
Automated systems that detect deception in \textit{high-stakes} situations can enhance societal well-being across medical, social work, and legal domains. Existing models for detecting high-stakes deception in videos have been \textit{supervised}, but labeled datasets to train models can rarely be collected for most real-world applications. To address this problem, we propose the first multimodal \textit{unsupervised} transfer learning approach that detects real-world, high-stakes deception in videos without using high-stakes labels. Our subspace-alignment (SA) approach adapts audio-visual representations of deception in lab-controlled \textit{low-stakes} scenarios to detect deception in real-world, \textit{high-stakes} situations. Our best unsupervised SA models outperform models without SA, outperform human ability, and perform comparably to a number of existing supervised models. Our research demonstrates the potential for introducing subspace-based transfer learning to model high-stakes deception and other social behaviors in real-world contexts with a scarcity of labeled behavioral data.
\end{abstract}
\begin{keywords}
transfer learning, deception detection 
\end{keywords}
\section{Introduction}
Advances in human-centered signal processing and multimodal machine learning are enabling the development of automated systems that can detect human social behaviors, including \textit{deception} \cite{10.1109/TPAMI.2018.2798607}. Deception involves the intentional communication of false or misleading information \cite{article} and has been categorized as occurring in either \textit{high-stakes} or \textit{low-stakes} social situations \cite{doi:10.1348/135532509X433151}. Deceivers in high-stakes contexts face substantial consequences if their deception is discovered, in contrast to deceivers in low-stakes contexts. Healthcare providers, social workers, and legal groups have fostered an interest in detecting \textit{high-stakes} deception for applications that enhance societal well-being (e.g., helping therapists recognize whether clients are masking negative emotions, helping judges assess courtroom testimonies of children coerced to lie) \cite{9053386, 10.1145/3107990.3108005}. Human deception detection ability has been determined as close to chance level \cite{Bond2006AccuracyOD}, motivating the development of computational approaches that can help humans in this challenging task. 

A fundamental challenge in modeling high-stakes deception is the \textit{scarcity of labeled high-stakes deception data}, due to the difficulty in collecting large amounts of real-world data with verifiable ground truth \cite{doi:10.1348/135532509X433151}. Lab-controlled experiments to simulate realistic high-stakes scenarios are not ethical, because they require the use of threats to impose substantial consequences on deceivers. Therefore, most researchers have collected data of participants communicating truthfully and deceptively in lab-controlled, low-stakes situations (e.g., mock crime scenarios) to study behavioral cues that could be indicative of high-stakes deception \cite{10.1145/3161178, doi:10.1348/135532509X433151}. 

Unsupervised domain adaptation (UDA) leverages knowledge from labeled \textit{source} domains to perform tasks in related, unlabeled \textit{target} domains \cite{5288526, 6751479}. We addressed the data scarcity problem of high-stakes deception by proposing a novel UDA approach, based on subspace alignment (SA) \cite{6751479}, to detect high-stakes deception without using any high-stakes labels. Grounded in psychology research regarding the generalizability of deceptive cues across contexts \cite{article}, \textit{we hypothesized that audio-visual representations of low-stakes deception in lab-controlled situations can be leveraged by SA to detect high-stakes deception in real-world situations}. 

We experimented with unimodal and multimodal audio-visual SA models to contribute effective modalities, fusion approaches, and behavioral signals for unsupervised high-stakes deception detection. Our best unsupervised SA models (75\% AUC, 74\% accuracy) outperform models without SA, outperform human ability, and perform comparably to a number of existing supervised models \cite{RillGarca2019HighLevelFF, 10.1145/2818346.2820758, 7836768, 10.1145/3349801.3349806}. Our research demonstrates the potential for introducing unsupervised subspace-based transfer learning approaches to address the scarcity of labeled data when modeling high-stakes deception and other social behaviors in real-world situations.

\section{Background}
Existing approaches for detecting high-stakes deception in videos have leveraged \textit{supervised} machine learning models that exploit discriminative patterns in human visual, verbal, vocal, and physiological cues to distinguish deceptive and truthful communication \cite{10.1145/3107990.3108005}. To address the scarcity of labeled high-stakes deception data for training models \cite{8621909}, prior research has focused on developing models that are \textit{robust} to small numbers of samples \cite{8621909, 8953413}. While \textit{supervised} transfer learning models have been developed to detect deception \cite{8925473, Luo2017}, to the best of our knowledge, no existing research has introduced \textit{unsupervised} models to address the data scarcity problem of high-stakes deception detection.

Psychology studies have found that humans can rely on audio-visual behavioral cues (e.g., facial expressions, voice pitch) to detect deception across different people and contexts in lab-controlled experiments  that attempt to simulate high-stakes situations (e.g., mock crime scenarios) \cite{article, Bond2006AccuracyOD}. These findings motivated our decision to leverage representations of audio-visual cues during low-stakes deception to detect high-stakes deception across different people and contexts. We note that subspace-based transfer learning has been successfully used to detect emotion \cite{yang2017weightedgeodesicflowkernel, 8276251}, but has not been previously developed to detect social behaviors. To the best of our knowledge, our unsupervised subspace-based transfer learning approach is the first to detect a social behavior, leveraging audio-visual cues to detect high-stakes deception. 

\section{Datasets}
This section describes the video datasets used for real-world \textit{high-stakes} deception and lab-collected \textit{low-stakes} deception. 

\subsection{High-stakes deception dataset}
For high-stakes deception, we used a publicly-available video dataset of people in 121 real-world courtroom trial situations ($\sim$28 sec per video); this dataset is the current benchmark for high-stakes deception detection in videos \cite{10.1145/2818346.2820758}. Each video was labeled as \textit{truthful} or \textit{deceptive} per police testimony and trial information; we used these labels as ground truth. Per criteria used by prior research with this dataset \cite{DBLP:conf/aaai/WuSDS18, 10.1145/3382507.3418864}, we identified a usable subset of 108 videos (53 truthful videos, 55 deceptive videos; 47 speakers of diverse race and gender). All videos were collected in unconstrained situations with a variety of illuminations, camera angles, and face occlusions.  

\subsection{Low-stakes deception dataset}
For low-stakes deception, we used the UR Lying Dataset \cite{10.1145/3161178}, the only publicly-available video dataset that had participants voluntarily communicating truthfully and deceptively in lab-controlled game scenarios ($\sim$23 min per video). Participants chose to respond either truthfully or deceptively to questions about images (e.g., picture of a flower). They won \$10 if they were believed while communicating truthfully and \$20 if they were believed while communicating deceptively. We used 107 videos (44 truthful videos, 63 deceptive videos; 29 speakers of diverse race and gender). All videos were collected in situations with consistent illuminations and camera angles, as well as minimal face occlusions.

\section{Methodology}
\begin{figure*}[t]
  \includegraphics[width=1\linewidth, height=5.9cm]{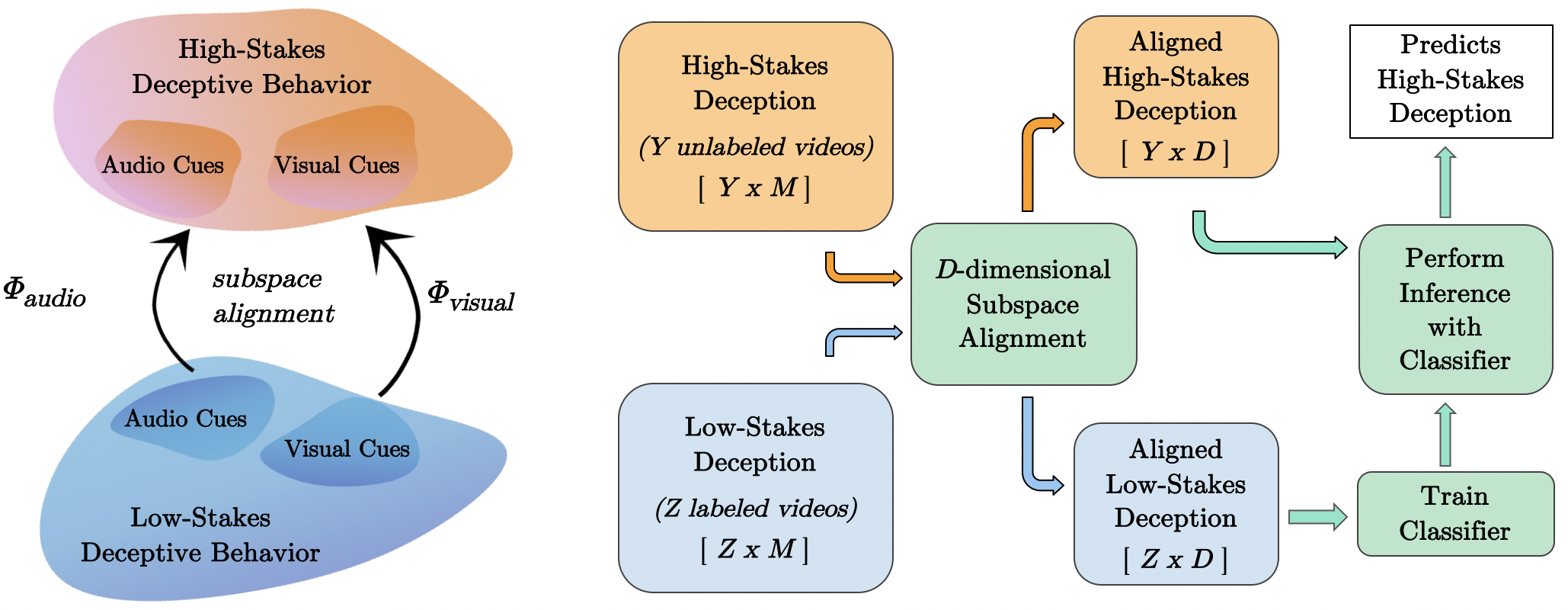}
  \caption{We propose \textit{unsupervised, audio-visual subspace alignment (SA)} for detecting high-stakes deception without using any high-stakes labels. \textit{M}-dimensional subspaces of low-stakes deception (blue) are aligned with high-stakes deception (orange) through transition matrices $\phi$ and used in our  \textit{D}-dimensional SA approach (green) to predict high-stakes deception.}
\end{figure*}
This section describes our approach (\textbf{Fig. 1}) for detecting \textit{high-stakes} deception through unsupervised audio-visual SA.
\subsection{Feature Extraction}
We extracted the same set of 89 audio-visual behavioral cues from speakers in both datasets. The OpenSMILE \cite{10.1145/2502081.2502224} toolkit was used to extract 58 audio features from the eGeMAPs \cite{7160715} and MFCC feature sets, to capture spectral, cepstral, prosodic, and voice quality information at each audio frame. The eGeMAPs and MFCC features sets in OpenSMILE overlap on MFCC features, and we do not include any duplicates. The OpenFace \cite{8373812} toolkit was used to extract 31 visual features to capture facial action units (FAU), head pose, and eye gaze information from speakers at each visual frame. 

The 89 audio-visual features were extracted frame-by-frame from videos. To prepare these features for a binary video classification task, we represented each as a fixed-length vector of time-series attributes during variable-length videos, similar to prior methods \cite{RillGarca2019HighLevelFF, 10.1145/3382507.3418864}. For each feature, the TsFresh toolkit \cite{tsfresh} computed 12 attributes: mean, standard deviation, aggregate autocorrelation across different time lags, and changes in feature values across quantiles. These functions are documented in the EfficientFCParameters class of TsFresh. A fixed-length feature vector of 1068 audio-visual time-series features (\textit{$89 \times 12$}) was computed to represent each video.      

\subsection{Unsupervised Subspace Alignment (SA)} 
We formulated high-stakes deception detection as an unsupervised subspace alignment (SA) transfer learning problem \cite{6751479}. Given a set \textit{S} of \textit{Z} labeled low-stakes deception samples (source domain) and a set \textit{T} of \textit{Y} unlabeled high-stakes deception samples (target domain), the modeling goal is to train a classifier on \textit{S} that predicts \textit{deceptive} and \textit{truthful} labels of \textit{T}. Both \textit{S} and \textit{T} are in \textit{M}-dimensional feature spaces and are drawn according to different marginal distributions. PCA computes the first \textit{$D$} principal components of each domain to create \textit{$D$}-dimensional subspace embeddings \textit{$C_{S}$} (source components) and \textit{$C_{T}$} (target components). The optimal linear transformation matrix \textit{$\phi$} is computed by minimizing the difference between \textit{$C_{S}$} and \textit{$C_{T}$} with the following equation:
\begin{equation} 
    \begin{split}
        \phi &= argmin_{\phi} ||C_{s}\phi - C_{T}||_{F}^{2} \\
        &= argmin_{\phi} ||C_{S}^{T}C_{S}\phi - C_{S}^{T}C_{T}||_{F}^{2}\\
        &= argmin_{\phi} ||\phi - C_{S}^{T}C_{T}||_{F}^{2}\\
        &= C_{S}^{T}C_{T}
    \end{split}
    \label{suba}
\end{equation}
where $||\cdot||$ denotes the Frobenius norm. This matrix $\phi$ is used to transform low-stakes deception embeddings \textit{$C_{S}$} to align with high-stakes deception. Classifiers are then trained on these aligned low-stakes embeddings and tested on high-stakes embeddings \textit{$C_{T}$} to predict \textit{deceptive} or \textit{truthful} labels for high-stakes samples. Our approach is visualized in \textbf{Fig. 1}; additional details on unsupervised SA are in \cite{6751479}.
\subsection{Classification Experiments}
We experimented with 9 different unimodal and multimodal unsupervised SA approaches, described below, to identify effective modalities, feature sets, and modeling approaches.
\subsubsection{Unimodal and Multimodal Classification Models}
Unimodal audio SA classifiers were trained on the set of all audio features and separately on MFCC and eGeMAPs feature sets. Unimodal visual SA classifiers were trained on the set of all visual features and separately on FAU, eye gaze, and head pose feature sets. Multimodal SA classifiers were trained with two approaches: (1) \textit{early-fusion} and (2) \textit{late-fusion}. Early-fusion SA classifiers were trained on concatenated feature vectors of all audio-visual features. For late-fusion, separate unimodal SA classifiers were trained on audio and visual features, and a majority vote of their predicted class probabilities determined the final prediction. The classifier for all SA experiments was K Nearest Neighbors (KNN), implemented with scikit-learn \cite{scikit-learn}. Similar to  KNN hyper-parameter tuning for SA in \cite{6751479}, all SA experiments were conducted with 3-fold cross-validation in the source subspace to identify optimal values of the KNN nearest-neighbors hyper-parameter \textit{k} in the discrete range [1, 30]. An optimal subspace dimension \textit{D} was determined in the discrete range [1, 10], since smaller values of \textit{D} avoid computationally expensive eigendecomposition.

\subsubsection{Baseline Models and Chance Performance}
For each unimodal and multimodal SA model, we implemented baseline KNN classifiers trained on low-stakes deception data and tested on high-stakes deception data, without alignment, in order to evaluate the effectiveness of SA. For fair comparison, each baseline KNN was implemented with the same hyper-parameter \textit{k} as the optimal \textit{k} that was automatically computed during 3-fold cross-validation in the corresponding SA model. To compare our model performance to human deception detection ability, which is at chance level \cite{Bond2006AccuracyOD}, we defined the human performance baseline as a classifier that would achieve 51\% accuracy (always predicting \textit{deceptive} for 55 deceptive videos out of 108 videos).
\subsubsection{Metrics}
Aligned with previous research \cite{10.1145/3382507.3418864, DBLP:conf/aaai/WuSDS18, RillGarca2019HighLevelFF}, the following metrics were computed to evaluate classifiers: (1) ACC, classification accuracy across the videos; (2) AUC, the probability of the classifier ranking a randomly chosen deceptive sample higher than a randomly chosen truthful one; (3) F1-score, weighted average of precision and recall. 
\begin{table}[]
\centering
\begin{tabular}{lccc}
\multicolumn{1}{c|}{\textbf{Model}}             & \multicolumn{1}{c|}{\textbf{ACC}}  & \multicolumn{1}{c|}{\textbf{AUC}}  & \textbf{F1}   \\ \hline
\multicolumn{4}{c}{\textbf{Audio}}                                                                                 \\ \hline
\multicolumn{1}{l|}{Audio (MFCC + eGeMAPs)}         & \multicolumn{1}{c|}{0.63} & \multicolumn{1}{c|}{0.65} & 0.69 \\
\multicolumn{1}{l|}{MFCC\textsuperscript{**}}                           & \multicolumn{1}{c|}{0.61} & \multicolumn{1}{c|}{0.60} & 0.62 \\
\multicolumn{1}{l|}{eGeMAPs}                        & \multicolumn{1}{c|}{0.55} & \multicolumn{1}{c|}{0.55} & 0.48 \\ \hline
\multicolumn{1}{l|}{Audio Baseline}                 & \multicolumn{1}{c|}{0.52} & \multicolumn{1}{c|}{0.48} & 0.60 \\
\multicolumn{1}{l|}{MFCC Baseline}                  & \multicolumn{1}{c|}{0.42} & \multicolumn{1}{c|}{0.42} & 0.53 \\
\multicolumn{1}{l|}{eGeMAPs Baseline}               & \multicolumn{1}{c|}{0.49} & \multicolumn{1}{c|}{0.50} & 0.35 \\ \hline
\multicolumn{4}{c}{\textbf{Visual}}                                                                                \\ \hline
\multicolumn{1}{l|}{Visual (FAU + Gaze + Pose)\textsuperscript{**}} & \multicolumn{1}{c|}{\textbf{0.74}} & \multicolumn{1}{c|}{\textbf{0.75}} & \textbf{0.73} \\
\multicolumn{1}{l|}{FAU\textsuperscript{*}}                            & \multicolumn{1}{c|}{0.64} & \multicolumn{1}{c|}{0.59} & 0.70 \\
\multicolumn{1}{l|}{Gaze\textsuperscript{*}}                           & \multicolumn{1}{c|}{0.70} & \multicolumn{1}{c|}{0.71} & 0.64 \\
\multicolumn{1}{l|}{Pose\textsuperscript{*}}                           & \multicolumn{1}{c|}{0.63} & \multicolumn{1}{c|}{0.62} & 0.61 \\ \hline
\multicolumn{1}{l|}{Visual Baseline}                & \multicolumn{1}{c|}{0.62} & \multicolumn{1}{c|}{0.63} & 0.59 \\
\multicolumn{1}{l|}{FAU Baseline}                   & \multicolumn{1}{c|}{0.51} & \multicolumn{1}{c|}{0.47} & 0.61 \\
\multicolumn{1}{l|}{Gaze Baseline}                  & \multicolumn{1}{c|}{0.52} & \multicolumn{1}{c|}{0.52} & 0.57 \\
\multicolumn{1}{l|}{Pose Baseline}                  & \multicolumn{1}{c|}{0.43} & \multicolumn{1}{c|}{0.42} & 0.52 \\ \hline
\multicolumn{4}{c}{\textbf{Audio-Visual}}                                                                          \\ \hline
\multicolumn{1}{l|}{Audio-Visual Early Fusion (EF)\textsuperscript{**}} & \multicolumn{1}{c|}{0.64} & \multicolumn{1}{c|}{0.60} & 0.69 \\
\multicolumn{1}{l|}{Audio-Visual Late Fusion (LF)}  & \multicolumn{1}{c|}{0.63} & \multicolumn{1}{c|}{0.65} & 0.68 \\ \hline
\multicolumn{1}{l|}{Audio-Visual EF Baseline}       & \multicolumn{1}{c|}{0.44} & \multicolumn{1}{c|}{0.44} & 0.53 \\
\multicolumn{1}{l|}{Audio-Visual LF Baseline}       & \multicolumn{1}{c|}{0.52} & \multicolumn{1}{c|}{0.47} & 0.66
\end{tabular}
\textsuperscript{*}\footnotesize{Significant difference ($p<0.05$) between this SA model and baseline}
\textsuperscript{**}\footnotesize{Significant difference ($p<0.01$) between this SA model and baseline}
\caption{Classification Results}
\end{table}
\raggedbottom
\section{Results and Discussion}
Modeling results from classification experiments are presented in \textbf{Table 1} and visualized in \textbf{Fig. 2}. All unimodal and multimodal unsupervised SA models substantially outperformed the baseline models without SA and the human performance chance level, \textit{demonstrating the effectiveness of unsupervised SA for modeling high-stakes deceptive behavior without any high-stakes labels}. Significance values of differences in model performance were computed with McNemar’s test ($\alpha$=0.05) with continuity correction \cite{mcnemar}.

The unimodal \textit{visual SA} model, trained on all visual features, had the highest performance (hyper-parameters \textit{D}=6 and \textit{k}=3). This model achieved an ACC of 74\%, AUC of 75\%, and F1-score of 73\%, and significantly outperformed the visual baseline model without SA (p$<$0.01). This model also outperformed the unimodal SA models trained on audio features and the multimodal early fusion and late fusion SA models trained on audio-visual features. Within the visual modality, SA models trained on the \textit{gaze} features outperformed SA models trained on the facial action units and head pose features. These results suggest that representations of visual behavior, in particular eye gaze, demonstrate more transfer potential than audio and audio-visual representations in subspace-based transfer learning approaches for detecting deception across different people in different social contexts. 

\begin{figure}[tp]
  \includegraphics[width=1\linewidth]{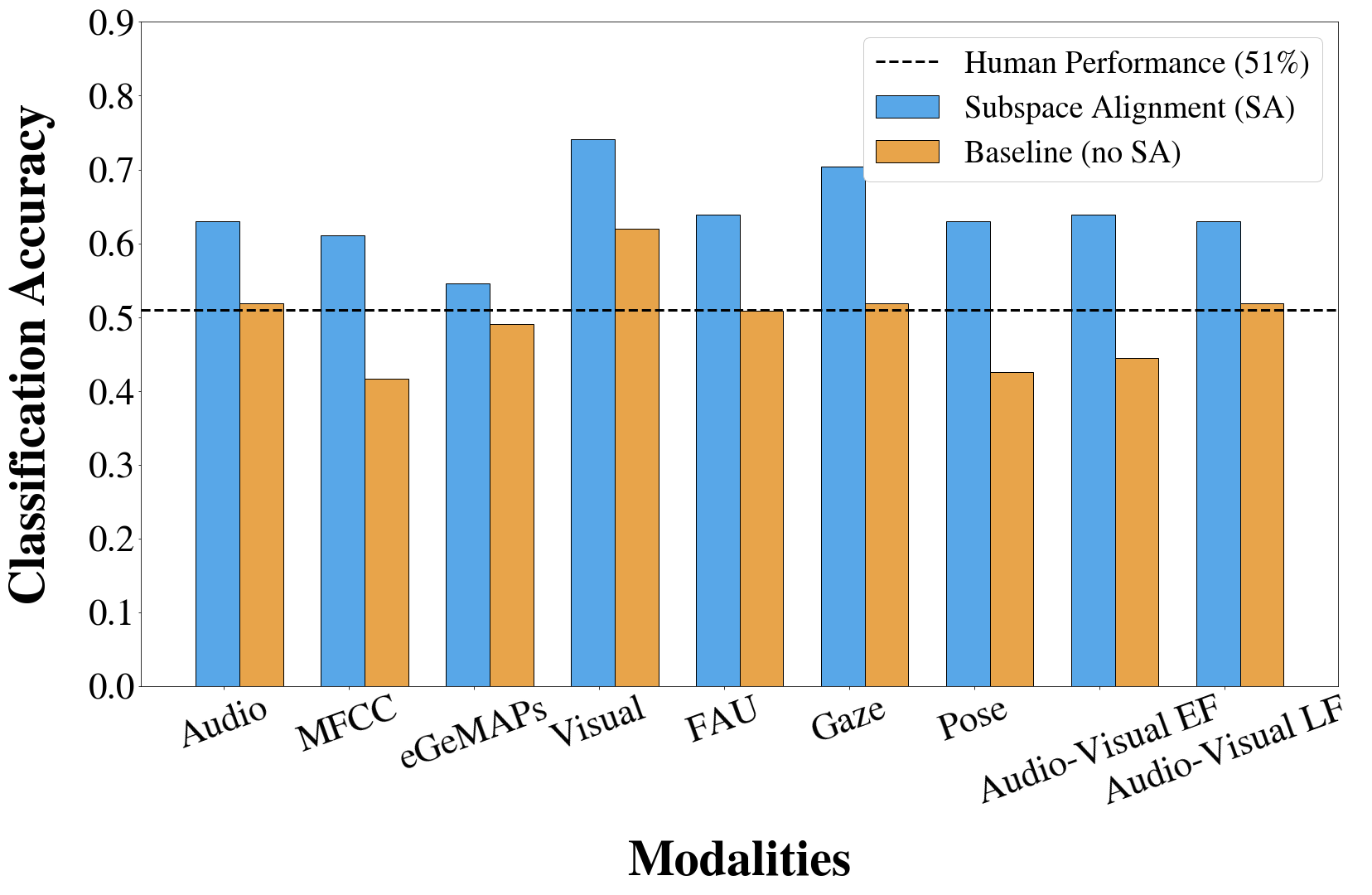}
  \caption{Classification accuracies of each unimodal and multimodal SA model (blue) compared to baseline models (orange) and the chance human performance (dashed line). }
\end{figure}

Our \textit{unsupervised} visual SA model performed comparably to existing \textit{fully-supervised}, automated approaches that used the same dataset (ACC 75\% \cite{10.1145/2818346.2820758}, 77\% \cite{10.1145/3349801.3349806}, 79\% \cite{7836768}; AUC 70\% \cite{RillGarca2019HighLevelFF}). While some prior fully-supervised, automated approaches \cite{10.1145/3382507.3418864, 8953413, DBLP:conf/aaai/WuSDS18} outperformed our unsupervised SA, our findings support the potential for introducing unsupervised SA to address the \textit{data scarcity} problem of modeling high-stakes deception. \textit{Our results support our hypothesis that audio-visual representations of low-stakes deception in lab-controlled situations can be leveraged by SA to detect high-stakes deception in real-world situations.} 

To analyze audio-visual behavioral signals that are similar in distribution across low-stakes and high-stakes deception contexts, we conducted a two-tail independent sample Welch’s t-test between each feature's low-stakes and high-stakes distributions without assuming equal variance \cite{welch}. The following 10 behavioral signals exhibited the most similarity in distribution: jitter, left eye gaze, pitch, yaw of head pose, roll of head pose, right eye gaze, the ratio of the first two harmonics of fundamental frequency, and the MFCC coefficients 0, 7, and 8. The similar distributions of these features across different groups of people and different social contexts demonstrate their potential for use as \textit{transferable} cues in models for cross-situational deception detection.   
\section{Conclusion}
This paper proposes the first \textit{unsupervised} transfer learning approach for detecting real-world, high-stakes deception in videos without using high-stakes labels. Our subspace-alignment models addressed the \textit{data scarcity problem} of modeling high-stakes deception by adapting audio-visual representations of deception in lab-controlled \textit{low-stakes} scenarios to detect deception in real-world, \textit{high-stakes} situations. We contribute effective modalities, modeling approaches, and audio-visual cues, for high-stakes deception detection. Our research demonstrates the potential for introducing subspace-based transfer learning approaches to model high-stakes deception and other social behaviors in real-world contexts, when faced with a scarcity of labeled behavioral data.
\section{Acknowledgements}
We thank the Rochester HCI Group for sharing their low-stakes deception dataset. This research was supported by the USC Provost’s Undergraduate Research Fellowship.
\bibliographystyle{IEEEbib}
\bibliography{strings,refs}
\end{document}